\documentclass[a4paper, 12pt]{article}

\usepackage{soul}
\usepackage{cite}

\usepackage{booktabs} 
\usepackage{colortbl} 
\usepackage{authblk}

\usepackage{etoolbox}
\usepackage{amsmath,amssymb,amsthm,amsfonts,mathtools}
\usepackage{xspace}
\usepackage[table]{xcolor}
\usepackage{dsfont}
\usepackage{bm}
\usepackage{subcaption}
\usepackage{hyperref}
\usepackage{tikz}
\usetikzlibrary{positioning}
\usepackage{pgfplots}
\usepgfplotslibrary{fillbetween}

\usepackage[algo2e,ruled,vlined,linesnumbered]{algorithm2e}
\usepackage{url}

\graphicspath{{pics/}}

\DeclareMathOperator{\Bin}{Bin}
\newcommand{\R}{\mathbb{R}}

\allowdisplaybreaks

\pgfplotsset{compat=newest}
\pgfplotscreateplotcyclelist{myplotcycle}{%
very thick,blue,dashed\\
very thick,blue\\
very thick,red!80!black,dashed\\
very thick,red!80!black\\
very thick,violet!80!black,dashed\\
very thick,violet!80!black\\
very thick,orange,dashed\\
very thick,orange\\
very thick,cyan!80!black,dashed\\
very thick,cyan!80!black\\
}

\pgfplotscreateplotcyclelist{myplotcycleA}{%
very thick,blue\\
thin,blue\\
thin,blue\\
}

\pgfplotscreateplotcyclelist{myplotcycleB}{%
very thick,violet!80!black\\
}

\pgfplotstableread[col sep=comma]{data/size-by-evals.csv}\sizebyevalstable
\pgfplotstableread[col sep=comma]{data/time-by-evals.csv}\timebyevalstable
\pgfplotstableread[col sep=comma]{data/time-by-size.csv}\timebysizetable

\begin{document}


\title{Improving Time and Memory Efficiency of Genetic Algorithms
by Storing Populations as Minimum Spanning Trees of Patches}

\author{Maxim Buzdalov}
\affil{Aberystwyth University, Aberystwyth, United Kingdom}

\maketitle

\begin{abstract}
In many applications of evolutionary algorithms the computational cost of applying operators
and storing populations is comparable to the cost of fitness evaluation. Furthermore, by knowing what
exactly has changed in an individual by an operator, it is possible to recompute fitness value much
more efficiently than from scratch. The associated time and memory improvements have been
available for simple evolutionary algorithms, few specific genetic algorithms and in the context of
gray-box optimization, but not for all algorithms, and the main reason is that it is difficult to
achieve in algorithms using large arbitrarily structured populations.

This paper makes a first step towards improving this situation. We show that storing the population as
a minimum spanning tree, where vertices correspond to individuals but only contain meta-information
about them, and edges store structural differences, or \emph{patches}, between the individuals,
is a viable alternative to the straightforward implementation. Our experiments suggest that
significant, even asymptotic, improvements~--- including execution of crossover operators!~---
can be achieved in terms of both memory usage and computational costs.
\end{abstract}

This is a slightly revised and extended author's version of the GECCO'23 paper
accepted to the EvoSoft workshop. That paper is available at DOI \href{https://doi.org/10.1145/3583133.3596388}{10.1145/3583133.3596388}.

\section{Introduction}
\label{sec:intro}

In evolutionary computation, the standard ways to measure the performance
of an evolutionary algorithm are either to count the number of fitness function evaluations
until a solution of a certain quality is found (the fixed-target perspective),
or to measure the quality of the best found solution once the given number of evaluations is used
(the fixed-budget perspective). Both implicitly assume that fitness evaluation dominates
the computational costs of running an evolutionary algorithm.

While this assumption often comes true, the opposite also happens quite often. For example,
many multiobjective algorithms, such as NSGA-II~\cite{nsga-ii}, and advanced continuous
evolutionary algorithms, such as CMA-ES~\cite{hansen-2001}, have internal state update routines
that dominate the computational costs asymptotically, so for large enough population size
fitness evaluation will become dominated by other parts of these algorithms. However, there are also
important cases where computational costs need to be considered accurately even for simple algorithms.

Auxiliary computational costs become especially noticeable if fitness evaluation takes time
proportional to the individual size and hence is cheap. This often happens in benchmarking of
evolutionary algorithms on problems with easy definitions: in discrete optimization,
examples are \textsc{OneMax} (the number of bits set to 1),
\textsc{LeadingOnes} (the length of the maximum prefix that has all bits set to 1),
and many continuous functions, starting from \textsc{Sphere} (the sum of squares for each variable),
are also computationally cheap. This holds true for many practical fitness functions,
such as those used for satisfiability problems~\cite{critical-behaviour-sat,whitley-chicano-goldman},
knapsack problems~\cite{chu-beasley-knapsacks,linear-knapsack-measures} 
and many other integer linear programming problems~\cite{ilp-next-release-problem}.
In this case, allocation of space for a new individual, as well as na{\"i}ve implementation of mutation
operators, require asymptotically the same number of operations as fitness evaluation, and,
depending on the hidden constants, they may even take more time.

Significant decrease in computational costs is possible whenever the fitness function
can be incrementally recomputed when the exact positions of changes are known. This is known as
\emph{partial evaluation} or \emph{incremental evaluation}. One prominent example is gray-box
optimization~\cite{whitley-chicano-goldman}, which benefits from incremental fitness evaluation
especially with large-scale problem sizes~\cite{deb-billion-variables}. Improvements can often be
applied to mutation
operators~\cite{chicano-moves-in-a-ball-gecco14,whitley-one-million-variables-nk-gecco17},
but fast specialized crossovers are also possible~\cite{whitley-gpx-tsp-ec2020}.
Even outside of gray-box optimization, many fitness functions, including the ones listed
in the previous paragraph, still enjoy large asymptotic runtime improvements.
To benefit from this, one should be really careful with operations happening inside
evolutionary algorithms, as even allocation of space for a new individiual becomes
prohibitively time- and memory-consuming.

As a result, efficient implementations are known for mutation-only algorithms, few genetic algorithms
with very specific crossover operators, and in the gray-box optimization contexts.
In the same time, most genetic algorithms having non-trivial populations and employing crossover operators
so far had no other choice than implementing most operators straightforwardly, lacking the ability
to save computational resources and in particular enjoy incremental fitness evaluation.
Most of this comes from the fact that there is no way to perform any kind of crossover
on individuals, that could have diverged long time ago, faster than by full evaluation in an absence
of an appropriate data structure. In this work, we show that it is, in fact, possible.

We propose to store the population as a graph where each individual corresponds to a vertex,
and edges store the differences, or \emph{patches}, between the individuals they connect.
At any time, there is only one individual that is stored as a whole, whereas for all other individuals
vertices store only the meta-information, such as the fitness value. To simplify the data structure,
and also to save as much space as possible, we use as the graph the minimum spanning tree, such that
the total size of all the patches is minimum possible to preserve connectivity. Should an individual
be used to produce an offspring, we \emph{promote} the only complete individual
to the corresponding vertex by applying all the patches on the way that leads to that vertex.
Similarly, we may compute the difference between individuals by \emph{combining} the patches
on the way that connect their vertices, which typically takes way less time than by scanning the
entire individuals. Then we can use this difference to apply the crossover operator to them,
which is, again, typically faster than doing it in a na{\"i}ve way.

As we only begin our explorations of this topic, we consider only bit strings as the search space,
and limit ourselves with a few simple algorithms, including, however, an algorithm with
a non-trivial population that is able to maintain some diversity in it. We consider two problems:
the benchmark problem \textsc{OneMax} and the knapsack problem, which both benefit from incremental
fitness evaluation. Our experiments show a remarkable improvement in both time and memory over the 
na{\"i}ve implementation: in most favorable conditions, the cost of one fitness evaluation becomes
$O(1)$ versus $O(n)$ for problem size $n$, but even in the presence of population diversity,
the average cost still appears to grow sublinearly.

\textbf{Structure of the paper}. Section~\ref{sec:related} introduces the employed definitions,
algorithms and problems, and also covers the related work on adjacent topics.
Section~\ref{sec:approach} explains the proposed approach, the minimum spanning tree of patches.
Section~\ref{sec:experiments} presents the results of experiments and their discussion.
Finally, Section~\ref{sec:conclusion} concludes and indicates numerous directions for future work.

\textbf{Availability of code and data}.
To adhere to reproducibility standards discussed in~\cite{reproducibility},
our code is available on GitHub\footnote{\url{https://github.com/mbuzdalov/patch-based-ga}. This version of the paper is accompanied by the release \href{https://github.com/mbuzdalov/patch-based-ga/releases/tag/v0.2}{v0.2}.}.

\section{Preliminaries and Related Work}
\label{sec:related}

In this work, we consider only algorithms working on bit strings, and problems defined on bit strings.
Though most of our ideas are applicable to more general classes of search spaces, and so are
most of considered evolutionary algorithms, we define them on bit strings for brevity.

We denote as $n$ the length of the bit string. The Greek letter $\mu$ typically defines
the parent population size, whereas $\lambda$ typically defines the number of offspring in one iteration.
The term ``u.a.r.'' stands for ``uniformly at random''. All optimizers are defined to run infinitely,
whereas termination conditions are considered external to optimizer definitions.

\textbf{Simple problems}. We consider a benchmark problem called \textsc{OneMax}, whose (simplified)
definition is as follows:
\begin{equation*}
    \textsc{OneMax}: \{0,1\}^n \to \R, x \mapsto |\{i \in [1..n] \mid x_i=1\}|.
\end{equation*}

This problem obviously benefits from incremental fitness evaluation: if one knows the
old fitness value, the new one can be determined by examining the changed positions only.

We also consider the \emph{knapsack problem}~\cite{pisinger}, one of the famous NP-hard problems.
Given $n$ items, the $i$-th of them has the weight $w_i$ and the value $v_i$,
and the knapsack capacity $W$, one has to find a selection of these items
with the maximum total value, such that their total weight does not exceed the capacity.
Denoting $\overline{w} = (w_i)_{i=1}^{n}$, $\overline{v} = (v_i)_{i=1}^n$,
for the purpose of this paper we define the genotype-to-phenotype mapping as follows:
\begin{equation*}
    K_{\overline{w}, \overline{v}, W}^{P}: \{0,1\}^n \to (\R, \R), \quad
        x \mapsto \left(\sum\nolimits_{i=1}^{n} x_i w_i, \sum\nolimits_{i=1}^{n} x_i v_i\right),
\end{equation*}
and the phenotype-to-fitness mapping as follows:
\begin{equation*}
    K_{\overline{w}, \overline{v}, W}^{F}: (\R, \R) \to \R, \quad
        (W_x, V_x) \mapsto \begin{cases}
            V_x, & \text{if } W_x \le W;\\
            -W_x, & \text{otherwise}.
        \end{cases}
\end{equation*}

In other words, the bit string of length $n$ defines which items are selected: bit value 1 in the $i$-th
position selects the $i$-th item. The fitness function favors increasing the total value as long as
all items fit the knapsack, and decreasing the total weight if they do not.

This two-level mapping is essential to incremental fitness evaluation, as the phenotypic pair consisting
of sums of weights and values can be easily recomputed on small changes, and the fitness value can follow
suit, whereas with the more direct approach one would have to recompute the fitness from scratch.
For simplicity, we consider the pair of sums as the fitness value, redefine
the comparison operation, and avoid the explicit concept of a phenotype.

\textbf{Randomized local search}. Perhaps the simplest algorithm to benefit both from efficient
implementation of operators and from incremental fitness evaluation is Randomized local search,
or RLS, whose pseudocode is presented in Algorithm~\ref{algo:rls}.

\begin{algorithm2e}[!t]
\caption{Randomized local search}\label{algo:rls}
\SetKwInOut{Input}{Input}
\Input{problem size $n$, maximized function $f: \{0,1\}^n \to \R$}
\textbf{Initialization:} Sample $x \in \{0,1\}^n$ u.a.r. and evaluate $f(x)$\;
\textbf{Optimization:}
\For{$t=1,2,3,\ldots$}{
Let $y$ be a copy of $x$ with one random bit flipped\;
Evaluate $f(y)$\;
\lIf{$f(y) \ge f(x)$}{$y \gets x$}
}
\end{algorithm2e}

Obviously, instead of creating the entire individual $y$ from scratch, one can just flip the $i$-th bit
in the parent, where $i$ is chosen u.a.r. randomly, then evaluate the changed individual, and if it is 
worse, then ``unflip'' that bit back. This way, the internal work for a single fitness evaluation,
other than the first one, is limited to $O(1)$ operations.

If incremental fitness evaluation is possible, we can definitely apply it.
For instance, for the knapsack problem the sum of weights is either increased
or decreased by the weight of the $i$-th item, depending on whether it was selected or not.
In this case, the total work for the single fitness evaluation needs only $O(1)$ operations.

\textbf{Standard bit mutation and the $(1+1)$ evolutionary algorithm}. For a minimalist
evolutionary algorithm capable of global search, the $(1+1)$ evolutionary algorithm,
or the $(1+1)$~EA, is often considered (Algorithm~\ref{algo:1p1}). It is very similar to
RLS, except that for the mutation operator it uses \emph{standard bit mutation} which,
with probability $p$, flips each bit independently.

\begin{algorithm2e}[!t]
\caption{The $(1+1)$ evolutionary algorithm}\label{algo:1p1}
\SetKwInOut{Input}{Input}
\Input{problem size $n$, mutation rate $p$, \newline maximized function $f: \{0,1\}^n \to \R$}
\textbf{Initialization:} Sample $x \in \{0,1\}^n$ u.a.r. and evaluate $f(x)$\;
\textbf{Optimization:}
\For{$t=1,2,3,\ldots$}{
Let $y$ be a copy of $x$ with each bit flipped independently of others with probability $p$\;
Evaluate $f(y)$\;
\lIf{$f(y) \ge f(x)$}{$y \gets x$}
}
\end{algorithm2e}

The general recommendation for a new problem, unless more knowledge is obtained, is to use $p=1/n$
such that $np = 1$ bit is flipped on average. The number of flipped bits follows
the well-known binomial distribution with parameters $n$ and $p$, which is reasonably well concentrated
around the value $np$. Under these conditions it is obvious that the $(1+1)$~EA also benefits a lot
from incremental fitness evaluation.

But is the mutation operator capable of delivering the matching performance?
Most people, including even many developers of well-known and widely-used software systems for evolutionary computation,
implement sampling from binomial distributions using the straightforward na{\"i}ve scheme,
such as the \texttt{BitFlipMutation}\footnote{Last revision at the submission time: \href{https://github.com/jMetal/jMetal/blob/f1a6d0ae6bb19885e491c84177f00a4965c96f17/jmetal-core/src/main/java/org/uma/jmetal/operator/mutation/impl/BitFlipMutation.java}{BitFlipMutation.java}}
class in jMetal~\cite{jmetal} or the \texttt{mutFlipBit}\footnote{Last revision at the submission time: \href{https://github.com/DEAP/deap/blob/cb20d979d3b62635cc330a9804aeb29523bffd42/deap/tools/mutation.py}{mutation.py}}
function in DEAP~\cite{deap}. However, the properties of a binomial distribution
allow sampling schemes of complexity $O(np)$, which matches the number of changes made to the individual.
This fact has been known in the evolutionary computation domain for a long time,
see~\cite[p.~238, Eq.~32.2]{back-evolutionary-computation-1} and,
apparently independently, \cite{jansenZ-algorithm-engineering}.
In fact, with some preprocessing, such sampling can be performed
faster~\cite{walker-constant-sampling-1,walker-constant-sampling-2,ziggurat,binomials-cacm,
binomials-hoermann,gentle-rng},
but since the number of changes to an individual is still $\Theta(np)$,
in this domain we do not benefit much from faster sampling anyway.

As a result, it is possible to run the $(1+1)$~EA with as few as $O(1)$ operations per fitness evaluation,
assuming the fitness function cooperates, but it requires some care from software developers.

\textbf{The $(1+(\lambda,\lambda))$ genetic algorithm}. Proposed a decade ago by the theoretic community,
this genetic algorithm has some unique properties with regards to certain function
classes~\cite{learning-from-black-box-thcs}.

Many of these properties come not just from using crossover,
but from the very specific kind of crossover. It takes a parent individual $x$ and an individual $y$
that has been generated by flipping $\ell$ bits chosen u.a.r. in $x$. Then it generates an offspring
by taking each bit from $y$ with probability $p_c$ and from $x$ otherwise, treating each bit independently.
In the default configuration, $\ell$ and $p_c$ are tied in such a way that $\ell \cdot p_c = \Theta(1)$
with large probability.

This particular crossover, as well as the overall structure of the algorithm, allows implementing it
efficiently much in the same manner as the algorithms above. As the number of bits $\ell$ is usually
much smaller than the problem size $n$, one can not just generate $y$, but also store the list of bits
that makes it different from the parent $x$. Then, crossover can be executed by just sampling
$\ell \cdot p_c$ bits from that list and flipping them in $x$. Although not always explicitly 
mentioned in the corresponding papers, this was the mechanism to enable experiments with problem sizes $n$
of order up to $10^6$ to $10^7$ in realistic times (few minutes per run) not just on problems
like \textsc{OneMax}, but on MAX-SAT problems as well~\cite{buzdalovD-gecco17-3cnf}, sometimes
even allowing sampling rather large values of $\ell$ from heavy-tailed
distributions~\cite{antipovBD-oll-heavy-tailed-algo22}.

Such crossover can be interpreted as mutation subsampling. While not making large difference on
bit strings, this consideration actually enabled porting this algorithm to other search spaces,
such as permutations~\cite{bassinB-gecco20-oll-perm}, with similar performance improvements
compared to na{\"i}ve implementations. However, the techniques employed to achieve these improvements
cannot be readily used with other algorithms.

\textbf{General form of ``good'' mutation and crossover operators}.
Evolutionary algorithms capable of optimizing arbitrary problems over certain search spaces
shall possess certain invariance properties in order to avoid preferring some (otherwise equivalent)
problem instances over other. For example, in continuous optimization, the obvious requirements
are translation invariance and scaling invariance, whereas many algorithms, such as CMA-ES,
also have rotational invariance (and hence invariance over general affine transformations).
Having such properties generally improves reliability of black-box optimization
algorithms~\cite{hansen-invariance-asoc11}, whereas algorithms pretending to be black-box but lacking
many of these properties are often used in fraudulent publications~\cite{sine-cosine-is-flawed}.

For the search space consisting of bit strings of a fixed length, there are two such properties:
invariance over flipping a bit (which is a rough equivalent of translation invariance)
and invariance over shuffling positions of bits. Algorithms that satisfy these properties are called
\emph{unbiased black-box optimization algorithms}~\cite{unbiased-bbc-algorithmica},
a similar concept exists for arbitrary discrete search spaces~\cite{generic-unbiased-algorithms}.
This concept can be extended to variation operators, such as mutation and crossover operators.
For mutation operators over bit strings,
any unbiased mutation operator can be represented as follows~\cite{unbiased-bbc-algorithmica}:
\begin{itemize}
    \item sample a number $\ell \in [0..n]$;
    \item when applied to an individual $x$, copy it and
          flip in the copy $\ell$ bits without replacement at positions sampled u.a.r.
\end{itemize}
For instance, the standard bit mutation with mutation rate $p$ samples $\ell$ from a binomial distribution
with parameters $n$ and $p$. Similarly, any crossover operator can be represented as
follows~\cite{doerr-johannsen-faster-blackbox,doerr2014reducing,
practice-aware-legacy,bulanovaB-multiple-offspring-gecco17}:
\begin{itemize}
    \item sample a function $f$ from $d \in [0..n]$ to
          a pair of numbers $\ell_d \in [0..d], \ell_s \in [0..n-d]$;
    \item when applied to two individuals $x_1$ and $x_2$, compute the number of differing bits $d$
          and then compute $\ell_d$ and $\ell_s$ using the function $f$;
    \item copy $x_1$ and flip in the copy $\ell_d$ bits at positions where $x_1$ and $x_2$ differ,
          and also flip $\ell_s$ bits at positions where $x_1$ and $x_2$ are same,
          both without replacement and sampled u.a.r.
\end{itemize}

Similar definitions exist for operators of higher arities, as well as for operators returning
multiple offspring~\cite{bulanovaB-multiple-offspring-gecco17}. This view enables a unified framework
and a single-entry implementation for support of arbitrary operators of a given arity.

\begin{algorithm2e}[!t]
\caption{The $(\mu+1)$ genetic algorithm}\label{algo:mp1}
\SetKwInOut{Input}{Input}
\Input{problem size $n$, mutation rate $p_m$, crossover probability $p_c$,
       maximized function $f: \{0,1\}^n \to \R$}
\textbf{Initialization:} Let the population $X = \emptyset$\;
\For{$i=1,2,\ldots,\mu$}{
Sample $x_i \in \{0,1\}^n$ u.a.r. and evaluate $f(x_i)$\;
Add $x_i$ to the population: $X \gets X \cup \{x_i\}$\;
}
\textbf{Optimization:}
\For{$t=1,2,3,\ldots$}{
Sample random number $c$\;
\If{$c < p_c$}{
    Sample $x$ from $X$ u.a.r., let $y$ be a copy of $x$
    with each bit flipped independently of others with probability $p_m$\;
}
\Else{
    Sample $x_1$, $x_2$ from $X$ u.a.r. with replacement\;
    Apply uniform crossover to $x_1$ and $x_2$, obtain $x'$:
    on each position, $x'$ takes a value from either $x_1$ or $x_2$ with probability 1/2\;
    Let $y$ be a copy of $x'$
    with each bit flipped independently of others with probability $p_m$\;
}
Evaluate $f(y)$\;
Add $y$ to the population: $X \gets X \cup \{ y \}$\;
Remove the individual with the smallest fitness from $X$, breaking ties u.a.r.\;
}
\end{algorithm2e}

\textbf{The $(\mu+1)$ genetic algorithm}. As an example of a genetic algorithm with a non-trivial
population that cannot be easily adapted to the use of high-performance variation operators we
consider the $(\mu+1)$ genetic algorithm, presented in Algorithm~\ref{algo:mp1}.
This is essentially a steady-state algorithm, where on each iteration a new individual is generated
either by only mutation, or by crossover followed by mutation, and then the worst individual is
removed from the population.

Note that uniform crossover, followed by standard bit mutation, can be expressed in the above-mentioned
framework as follows: if the distance between the parents is $d$, then sample the number of differing
bits to flip $\ell_d$ from the binomial distribution $\Bin(d, 1/2)$, and the number of same bits to flip
$\ell_s$ from the binomial distribution $\Bin(n-d, p_m)$.

Despite its simplicity, the $(\mu+1)$~GA can be shown to be provably better than
mutation-only algorithms even on simple problems~\cite{pinto-doerr-crossover-ppsn18}, furthermore
proofs exist that increasing the population size improves the
performance~\cite{mu-plus-one-ga-populations}. With only a simple additional diversity-preserving
mechanism it becomes capable of efficiently jumping over large valleys~\cite{DangFKKLOSS18},
and even in its original form it is capable of preserving enough diversity to do essentially the
same~\cite{mu-plus-one-lasting-diversity-arxiv}.

Potential diversity of the population makes it a problem to perform crossovers efficiently,
as in the absence of any extra information the distance between the individuals
is essentially unknown and can change a lot over the run. To the best of our knowledge,
this paper is the first to present a mechanism that is capable of performing crossovers in
sublinear time in practice.

\textbf{Gray-box optimization}. This is the prominent area of research where, based on a deeper
understanding of the problem, efficient operators are possible to design with significant improvement in
in computational time, search efficiency, or both. A large number of works essentially belonging
to gray-box optimization implement mutation operators by essentially synthesizing the difference
vectors and storing only a few complete
individuals~\cite{deb-billion-variables,bosman-partial-evaluations-ec2021,
bosman-linkage-million-variables-gecco17}. These works either use only mutation operators or
invoke crossover operators only infrequently to amortize their cost.

With more understanding of the structure of the problem,
fast determininstic mutation operators and linear-time deterministic recombination that synthesizes
the best out of $2^q$ individuals from two individuals whose difference is partitioned into $q$
independent regions, is possible~\cite{whitley-one-million-variables-nk-gecco17,whitley-chicano-goldman,
whitley-gpx-tsp-ec2020,whitley-nk-clustering}, which leads to state-of-the-art solutions in
the particular problem domains. However, these advanced techniques cannot be used when the problem
admits incremental fitness evaluation, but the degree of linkage between variables is too high,
of which the knapsack problem and its variations are good examples.

\textbf{Individuals as immutable balanced trees}.
An approach that targets essentially the same problem, but in a significantly different way,
was recently proposed in~\cite{pitzerA-logarithmic-fitness-cec2021}.
For a class of problems where it is not only possible to perform incremental fitness evaluation,
but also to store the partial evaluation results for parts of individuals, optimization can benefit
by representing individuals as immutable balanced trees that store evaluation results for subtrees.
Note that both \textsc{OneMax} and the knapsack problem, which we consider in this paper,
belong to this class, but, for instance, most satisfiability problems do not.

In the case of bit strings, a mutation of $O(1)$ bits
corresponds to $O(\log n)$ time and memory to construct a new individual and compute its fitness.
What is more, the single-point crossover
(and any similar crossover with a constant number of exchange points)
can also be performed in time and memory of $O(\log n)$.
Note that such crossover operators are not unbiased, and direct implementation of, say, uniform
crossover using this approach does not result in any performance improvements. However, 
further development of this approach, such as aggressive subtree caching,
can probably make it more similar to our approach in many aspects.


\section{Minimum Spanning Tree of Patches}
\label{sec:approach}

Our data structure to work with the population consists of two parts:
the persistent lightweight graph and the mutable memory-heavy part of size $\Theta(n)$.

\textbf{The graph}. We maintain the entire population as a graph
that describes the structure of the population.
\begin{itemize}
    \item Vertices of this graph represent individuals. Only a limited amount of information
          about an individual is permanently stored in the vertex. Currently, this information
          consists of the fitness value of the individual, and a Boolean value which indicates
          whether an individual is still in the population (that is, it has not been discarded).
    \item Edges represent differences between individuals, or \emph{patches}. An edge from vertex
          $A$ to vertex $B$ contains an immutable piece of information that tells how to convert
          individual $A$ to individual $B$. In the general case, the graph is directed, and an edge
          from $B$ to $A$ shall produce an opposite effect to an edge from $A$ to $B$. However,
          for bit strings, these effects coincide: the edges contain a list of bit indices to flip.
\end{itemize}

This graph is maintained in such a way that it is a minimum spanning tree, where the weight of an edge
is the size of the patch (that is, the number of indices to flip). As even in the general case
it is reasonable to expect that the size of the patch is equal to the size of its negation, for the
purpose of building the tree we can safely use the characteristic size of a patch for a weight.

\textbf{The memory-heavy part}.
Apart from this graph, there are two more values which use $\Theta(n)$ memory. The first of these values
is what we call the \emph{complete individual}, which is a bit string of size $n$ containing one of
the individuals currently in the population. We also maintain a pointer to the vertex that corresponds
to this individual. The second value is what we call the \emph{mutable patch}: as opposed to immutable
patches stored at the edges, the mutable patch has the size of $\Theta(n)$ words. Its purpose is
to store a set of (zero-based) indices, which is a subset of $[0..n-1]$,
with the following supported operations:
\begin{enumerate}
    \item Clear the set.\label{op:clear}
    \item Add or remove a given element.\label{op:add}
    \item Test whether a given element is contained in the set.\label{op:test}
    \item Report the number of elements currently in the set.\label{op:size}
    \item Add a random element which is not yet in the set.\label{op:mut}
    \item Add a given number of random elements not yet in the set while
          \emph{simultaneously} removing another given number of random elements
          which are already in the set.\label{op:cross}
    \item Create an immutable set containing the same elements.\label{op:apply}
\end{enumerate}

The mutable patch can be implemented with some minimal care atop of one permutation $\pi$
of the set $[0..n-1]$, its inverse, and the size of the patch $s$, such that elements
$\{\pi(0), \ldots, \pi(s - 1)\}$ are the elements constituting the patch. For further details,
please refer to the class \texttt{MutableIntSet} in the repository.

Using operations (\ref{op:add})--(\ref{op:size}), we can combine in-place the mutable patch
and one of the immutable patches. This allows, in particular,
to traverse the entire graph using depth-first search (in time proportional to the sum of sizes
of all patches) and to find out which vertex results in the smallest cardinality of the set.
When adding a new individual to the population, this is used to find the existing individual
with which this new individual will be connected by an edge to ensure that the graph is the minimum
spanning tree. Similarly, by starting a depth-first search from the first parent and terminating it
when the second parent is encountered allows to compute the difference between the parents,
which will be used later in the crossover operator.

Operation (\ref{op:mut}) is used to implement mutation: given the number of bits to flip,
we first clear the set, then add the necessary number of not yet existing elements.
Similarly, operation (\ref{op:cross}) is used to implement crossover: assuming the set contains
the bit indices in which the parents differ, we ``flip'' the given number of randomly chosen bits
at positions where the parents differ and where they are the same, and after that the set
will contain exactly those bit indices which have to be flipped in the first parent.
For both operators, operation (\ref{op:apply}) is used to create a new immutable patch from the
result of applying the operator, that will then be used either to generate a new individual
or to create a permanent link from that individual to the existing population.

\begin{figure}[!t]
\centering
\begin{tikzpicture}[scale=1.3,
                    vertex/.style={circle, draw, minimum width=5mm},
                    empty/.style={vertex, fill=pink},
                    busy/.style={vertex, fill=yellow},
                    e/.style={anchor=south, sloped},
                    s/.style={rectangle, draw, fill=yellow}]
    \node[busy] (n0) at (0,2) {4};
    \node[empty] (n1) at (2,2) {5};
    \node[empty] (n2) at (4,2) {3};
    \node[empty] (n3) at (4,1) {1};
    \node[empty] (n4) at (6,1.5) {0};

    \draw (n0) edge node[e] {$\{0,3\}$} (n1);
    \draw (n1) edge node[e] {$\{1\}$}   (n2);
    \draw (n1) edge node[e] {$\{2\}$}   (n3);
    \draw (n3) edge node[e] {$\{1,3\}$} (n4);

    \node[s] (s0) at (0,0.2) {1};
    \node[s,right=0pt of s0] (s1) {1};
    \node[s,right=0pt of s1] (s2) {0};
    \node[s,right=0pt of s2] (s3) {1};
    \node[s,right=0pt of s3] (s4) {0};
    \node[s,right=0pt of s4] (s5) {0};
    \node[s,right=0pt of s5] (s6) {1};
    \node[s,right=0pt of s6] (s7) {1};
    \node[s,right=0pt of s7] (s8) {0};
    \node[s,right=0pt of s8] (s9) {1};
    \node[s,right=0pt of s9] (s10) {1};

    \draw[->, thick, dashed] (s5.north)--(n0);
\end{tikzpicture}
\par (a) Initial state \par
\vspace{3ex}
\begin{tikzpicture}[scale=1.3,
                    vertex/.style={circle, draw, minimum width=5mm},
                    empty/.style={vertex, fill=pink},
                    busy/.style={vertex, fill=yellow},
                    e/.style={anchor=south, sloped},
                    s/.style={rectangle, draw, fill=yellow},
                    ch/.style={s,fill=cyan}]
    \node[empty] (n0) at (0,2) {4};
    \node[busy] (n1) at (2,2) {5};
    \node[empty] (n2) at (4,2) {3};
    \node[empty] (n3) at (4,1) {1};
    \node[empty] (n4) at (6,1.5) {0};

    \draw (n0) edge node[e] {$\{0,3\}$} (n1);
    \draw (n1) edge node[e] {$\{1\}$}   (n2);
    \draw (n1) edge node[e] {$\{2\}$}   (n3);
    \draw (n3) edge node[e] {$\{1,3\}$} (n4);

    \node[ch] (s0) at (0,0.2) {0};
    \node[s,right=0pt of s0] (s1) {1};
    \node[s,right=0pt of s1] (s2) {0};
    \node[ch,right=0pt of s2] (s3) {0};
    \node[s,right=0pt of s3] (s4) {0};
    \node[s,right=0pt of s4] (s5) {0};
    \node[s,right=0pt of s5] (s6) {1};
    \node[s,right=0pt of s6] (s7) {1};
    \node[s,right=0pt of s7] (s8) {0};
    \node[s,right=0pt of s8] (s9) {1};
    \node[s,right=0pt of s9] (s10) {1};

    \draw[->, thick, dashed] (s5.north)--(n1);
\end{tikzpicture}
\par (b) The complete individual is promoted to vertex with fitness 5 \par
\vspace{3ex}
\begin{tikzpicture}[scale=1.3,
                    vertex/.style={circle, draw, minimum width=5mm},
                    empty/.style={vertex, fill=pink},
                    busy/.style={vertex, fill=yellow},
                    e/.style={anchor=south, sloped},
                    s/.style={rectangle, draw, fill=yellow},
                    ch/.style={s,fill=cyan}]
    \node[empty] (n0) at (0,2) {4};
    \node[empty] (n1) at (2,2) {5};
    \node[empty] (n2) at (4,2) {3};
    \node[empty] (n3) at (4,1) {1};
    \node[empty] (n4) at (6,1.5) {0};
    \node[busy] (n5) at (0,1) {6};

    \draw (n0) edge node[e] {$\{0,3\}$} (n1);
    \draw (n1) edge node[e] {$\{1\}$}   (n2);
    \draw (n1) edge node[e] {$\{2\}$}   (n3);
    \draw (n3) edge node[e] {$\{1,3\}$} (n4);
    \draw (n1) edge[dashed] node[e] {$\{2,9\}$}   (n5);

    \node[s] (s0) at (0,0.2) {0};
    \node[s,right=0pt of s0] (s1) {1};
    \node[ch,right=0pt of s1] (s2) {1};
    \node[s,right=0pt of s2] (s3) {0};
    \node[s,right=0pt of s3] (s4) {0};
    \node[s,right=0pt of s4] (s5) {0};
    \node[s,right=0pt of s5] (s6) {1};
    \node[s,right=0pt of s6] (s7) {1};
    \node[s,right=0pt of s7] (s8) {0};
    \node[ch,right=0pt of s8] (s9) {0};
    \node[s,right=0pt of s9] (s10) {1};

    \draw[->, thick, dashed] (s5.north)--(n5);
\end{tikzpicture}
\par (c) A new offspring is created \par
\vspace{3ex}
\begin{tikzpicture}[scale=1.3,
                    vertex/.style={circle, draw, minimum width=5mm},
                    empty/.style={vertex, fill=pink},
                    busy/.style={vertex, fill=yellow},
                    e/.style={anchor=south, sloped},
                    s/.style={rectangle, draw, fill=yellow},
                    ch/.style={s,fill=cyan}]
    \node[empty] (n0) at (0,2) {4};
    \node[empty] (n1) at (2,2) {5};
    \node[empty] (n2) at (4,2) {3};
    \node[empty] (n3) at (4,1) {1};
    \node[empty] (n4) at (6,1.5) {0};
    \node[busy] (n5) at (0,1) {6};

    \draw (n0) edge node[e] {$\{0,3\}$} (n1);
    \draw (n1) edge node[e] {$\{1\}$}   (n2);
    \draw (n1) edge node[e] {$\{2\}$}   (n3);
    \draw (n3) edge node[e] {$\{1,3\}$} (n4);
    \draw (n3) edge node[e] {$\{9\}$}   (n5);

    \node[s] (s0) at (0,0.2) {0};
    \node[s,right=0pt of s0] (s1) {1};
    \node[s,right=0pt of s1] (s2) {1};
    \node[s,right=0pt of s2] (s3) {0};
    \node[s,right=0pt of s3] (s4) {0};
    \node[s,right=0pt of s4] (s5) {0};
    \node[s,right=0pt of s5] (s6) {1};
    \node[s,right=0pt of s6] (s7) {1};
    \node[s,right=0pt of s7] (s8) {0};
    \node[ch,right=0pt of s8] (s9) {0};
    \node[s,right=0pt of s9] (s10) {1};

    \draw[->, thick, dashed] (s5.north)--(n5);
\end{tikzpicture}
\par (d) A patch of the minimum size is found for the new offspring \par
\caption{Minimum spanning tree of patches. Vertices contain fitness values,
         edges are labeled with patches: the sets of bit indices to flip.
         An example of performing a mutation operator on an individual with
         fitness 5 is shown}\label{fig:mst-mut}
\end{figure}

\textbf{The operations}. The following operations are supported.
\begin{itemize}
    \item Create a random individual. The first such call creates the memory-heavy part,
          initializes the \emph{complete individual} randomly,
          and creates the first vertex in the graph.
          All subsequent calls effectively perform mutation with $\ell \sim \Bin(n,1/2)$,
          such that the offspring is statistically identical to a randomly generated
          individual.
    \item Mutate $\ell$ bits in the given individual $x$. First, the \emph{complete individual}
          is promoted to match the individual $x$, which is performed by the depth-first search
          from $x$ until the vertex matching the complete individual is found, then by returning
          back and applying patches from the reverse edges. Then, the mutable patch is initialized
          by adding $\ell$ random elements that are not yet added. Next, starting from
          this state of the mutable patch, the entire graph is traversed completely
          in order to update it to reflect the new minimum spanning tree. When an edge is traversed,
          the patch written on that edge is prepended to the mutable patch, and the edge length
          between the current vertex and the new individual is the size of the mutable patch.
          The graph is rebuilt to become the new minimum spanning tree by an adaptation
          of an algorithm from~\cite{mst-updates-1978}, which works in linear time. When this
          algorithm ends, the shortest edge from an existing individual to the newly created
          individual is used to compute the fitness of the latter incrementally.
          An example application of the mutation operator is given in Fig.~\ref{fig:mst-mut}.
    \item Apply crossover to the given parents $x_1$ and $x_2$ using a function $f$ to obtain
          the numbers of bits to flip in the ``same'' and ``differing'' bit groups.
          First, the \emph{complete individual} is promoted to match $x_1$,
          and then the mutable patch is set to the difference between $x_1$ and $x_2$
          using depth-first search started from $x_1$ until $x_2$ is encountered.
          Then, the distance between the parents is obtained from the mutable patch,
          the function $f$ is called, and the simultaneous flip is called on the mutable patch.
          Finally, the procedure similar to the one in the mutation operator is followed
          to create the new vertex and connect it to the rest of the graph.
    \item Discard an individual. The individual is marked as discarded, and if it has degree 1 in the
          graph, it is removed from the graph. Then, if the former neighbor has also been marked as
          discarded, the same procedure is called recursively. In the operations above, when
          computing the shortest distance and finding an individual with that distance, only
          vertices not marked as discarded are used.
\end{itemize}

All these operations traverse the entire graph at least once (for discarding an individual, this is
only the worst-case bound), which takes time proportional to the sum of sizes of all the patches in
the graph (each individual creation also adds $\Theta(n)$ time and memory).
This total size is small for all the algorithms which are known to benefit from working with
patches and from incremental fitness evaluation, and can potentially be large
for large diverse populations. However, when an optimizer has worked for some time, even if it maintains
diversity, it may happen that a large portion of bit indices has converged to the same values
for the entire population, just because all other values are strictly worse. The proposed data structure
will automatically benefit from such conditions, as the largest patch sizes, and hence the average
distances between the individuals, will get smaller over time.

To measure the potential speed-ups from using the minimal spanning tree of patches, we conducted
a series of experiments, which we report in the next section.


\section{Experiments}
\label{sec:experiments}

In our experiments, we use the algorithms mentioned in Section~\ref{sec:related}
with the following notation used:
\begin{itemize}
    \item RLS: Randomized local search;
    \item $(1+1)$: the $(1+1)$ evolutionary algorithm
          using standard bit mutation with flip probability $1/n$;
    \item $(2+1)$: the $(\mu+1)$ genetic algorithm with $\mu=2$
          using standard bit mutation with flip probability $1.2/n$
          as suggested in~\cite{oliveto-2p1ga-tight-bounds}
          and crossover probability $0.9$;
    \item $(10+1)$: the $(\mu+1)$ genetic algorithm with $\mu=10$
          using standard bit mutation with flip probability $1.4/n$
          as suggested in~\cite{mu-plus-one-ga-populations}
          and crossover probability $0.9$.
\end{itemize}

For the $(\mu+1)$ genetic algorithm, we use different population sizes to see how they
affect the performance of the proposed data structure. All these algorithms are evaluated
using both the na{\"i}ve approach of handling populations and the proposed data structure.
In all cases, mutation operators using binomial distributions are implemented using efficient approaches.

As the project is implemented in Scala~3, which uses a Java virtual machine, to measure the running times
we need to warm up the virtual machine, which we do by observing the reported running times
over periods of one second (or more if the optimizer runs longer) and normalizing them over the
number of fitness evaluations. The obtained values, with a semantic of average operation time,
are remarkably stable, so we report their means and standard deviations over 10 such one-second attempts.

To perform the experiments, we use a dedicated Linux server with kernel 6.1.19,
OpenJDK Java virtual machine version 11.0.18, Scala version 3.2.2,
running on an Intel Core~i7-8700 CPU clocked at 3.2GHz.

\subsection{OneMax}

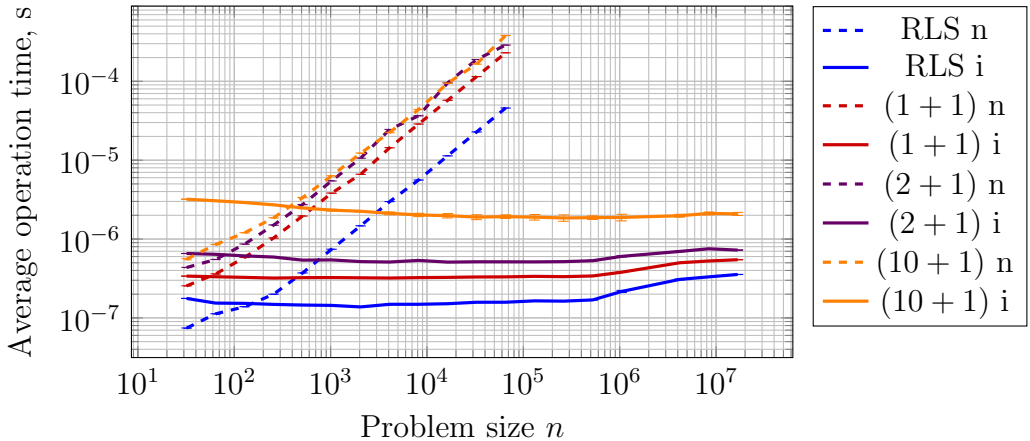
\begin{figure}[!t]
\centering
\begin{tikzpicture}
\begin{axis}[width=0.75\columnwidth, height=0.3\textheight, legend pos=outer north east,
             xmode=log, ymode=log, cycle list name=myplotcycle, grid=both,
             xlabel={Problem size $n$}, ylabel={Average operation time, s}]
    \addplot+ plot[error bars/.cd, y dir=both, y explicit]
              coordinates{
                (32,7.421305655467104E-8)+-(0,3.2982599173201406E-10)(64,1.1252236734672444E-7)+-(0,3.6265347212904046E-10)(128,1.38545682124874E-7)+-(0,4.906023936407037E-10)
                (256,1.9976067340427428E-7)+-(0,1.5498019922160904E-10)(512,3.717921711777774E-7)+-(0,2.707228723263824E-10)(1024,7.426736874873416E-7)+-(0,5.625593675005728E-10)
                (2048,1.4701183041956155E-6)+-(0,5.55530638410474E-9)(4096,2.983689578849757E-6)+-(0,1.6362778650027394E-9)(8192,5.605879157625321E-6)+-(0,2.6677736977961507E-9)
                (16384,1.1345617557163614E-5)+-(0,2.3914138506630168E-8)(32768,2.2813421998662258E-5)+-(0,1.6577224296311933E-8)(65536,4.583644555763083E-5)+-(0,8.616774484812065E-9)
              };
    \addlegendentry{RLS n};
    \addplot+ plot[error bars/.cd, y dir=both, y explicit]
              coordinates{
                (32,1.7664979156281186E-7)+-(0,5.0542421433248986E-11)(64,1.5411422050713973E-7)+-(0,1.6197917207000272E-10)(128,1.5292467905768935E-7)+-(0,7.631226358693081E-11)
                (256,1.4838327542339496E-7)+-(0,1.1177368710048617E-10)(512,1.45600983219058E-7)+-(0,3.9036776551148815E-10)(1024,1.4409389552645544E-7)+-(0,9.25916735237846E-11)
                (2048,1.378627621443594E-7)+-(0,2.657262760354466E-10)(4096,1.4878307856017667E-7)+-(0,6.22138988486762E-11)(8192,1.4887438903730523E-7)+-(0,7.04666052331481E-11)
                (16384,1.513429662565886E-7)+-(0,9.916221978833446E-11)(32768,1.579977370569038E-7)+-(0,1.1790623551654411E-10)(65536,1.5806108421626627E-7)+-(0,1.0674921528781138E-10)
                (131072,1.6456681440563059E-7)+-(0,4.670077836012331E-10)(262144,1.6312515070886532E-7)+-(0,2.447066120570977E-10)(524288,1.690962608680338E-7)+-(0,1.2674632840928396E-9)
                (1048576,2.165318150881107E-7)+-(0,6.80767622044366E-9)(4194304,3.078341090310793E-7)+-(0,3.598491246312858E-9)(8388608,3.3138698297273863E-7)+-(0,2.578760519648967E-9)
                (16777216,3.555746025157278E-7)+-(0,9.137172732616452E-10)
              };
    \addlegendentry{RLS i};
    \addplot+ plot[error bars/.cd, y dir=both, y explicit]
              coordinates{
                (32,2.548993237641765E-7)+-(0,8.144492822649103E-10)(64,3.6022656952748854E-7)+-(0,1.9589552220437E-9)(128,5.837305429601349E-7)+-(0,2.1360140708987347E-9)
                (256,1.031689082546112E-6)+-(0,4.451491732016482E-9)(512,1.9597527999304045E-6)+-(0,5.7740301763830685E-9)(1024,3.816573178106269E-6)+-(0,1.5602225796692577E-8)
                (2048,6.617333479371744E-6)+-(0,1.5047301935495936E-8)(4096,1.436833645890158E-5)+-(0,5.805056868748367E-8)(8192,2.8704204540658863E-5)+-(0,7.181259403666602E-8)
                (16384,5.737996075550658E-5)+-(0,6.778871935741194E-8)(32768,1.1478966467124681E-4)+-(0,9.845477779283982E-8)(65536,2.29733013533805E-4)+-(0,9.36107014314207E-8)
              };
    \addlegendentry{$(1+1)$ n};
    \addplot+ plot[error bars/.cd, y dir=both, y explicit]
              coordinates{
                (32,3.387498887557235E-7)+-(0,2.441626502010658E-10)(64,3.327532280334156E-7)+-(0,2.703401073049104E-10)(128,3.265822719595831E-7)+-(0,8.872098032878279E-10)
                (256,3.200327201359535E-7)+-(0,1.5844315819304446E-10)(512,3.23778657869973E-7)+-(0,5.457467158548083E-10)(1024,3.2472570984634613E-7)+-(0,1.809089907486856E-10)
                (2048,3.225796096097061E-7)+-(0,1.0009578670184532E-9)(4096,3.2031440522430267E-7)+-(0,2.3430021721435876E-10)(8192,3.2427552642694776E-7)+-(0,2.3085363117131814E-10)
                (16384,3.2650782751047783E-7)+-(0,5.104656119044572E-10)(32768,3.3039123063351277E-7)+-(0,1.5593460497494902E-9)(65536,3.313330624599161E-7)+-(0,1.4022966361577205E-10)
                (131072,3.365876032380258E-7)+-(0,8.489939565205685E-10)(262144,3.3337855571567893E-7)+-(0,1.4291704243105175E-9)(524288,3.4021510300413266E-7)+-(0,3.835236703738461E-9)
                (1048576,3.799098642245975E-7)+-(0,6.651288878288943E-10)(4194304,5.002502015842802E-7)+-(0,2.7894710416092017E-9)(8388608,5.268979152173791E-7)+-(0,1.598503536287015E-9)
                (16777216,5.465984830670781E-7)+-(0,2.1663792189362267E-9)
              };
    \addlegendentry{$(1+1)$ i};
    \addplot+ plot[error bars/.cd, y dir=both, y explicit]
              coordinates{
                (32,4.3567370480085337E-7)+-(0,1.3989056670994904E-9)(64,5.503264206837037E-7)+-(0,1.755677007877402E-9)(128,8.621639218318788E-7)+-(0,3.0914860487896247E-9)
                (256,1.5031216737675715E-6)+-(0,8.292890925933816E-9)(512,2.735059635467054E-6)+-(0,1.0868777376383298E-8)(1024,5.468904098418434E-6)+-(0,2.2613408123840465E-8)
                (2048,1.0624274692001283E-5)+-(0,3.7196136489865155E-8)(4096,2.44854069261609E-5)+-(0,6.522567610880668E-8)(8192,3.659567778969365E-5)+-(0,8.037093477452834E-8)
                (16384,9.541599719813366E-5)+-(0,1.3406876644477028E-7)(32768,1.8872371493129409E-4)+-(0,4.1196251193675565E-7)(65536,2.881574482605874E-4)+-(0,2.4922859122244995E-7)
              };
    \addlegendentry{$(2+1)$ n};
    \addplot+ plot[error bars/.cd, y dir=both, y explicit]
              coordinates{
                (32,6.557485573544221E-7)+-(0,1.327404704880916E-9)(64,6.402981252546445E-7)+-(0,2.0161928113522866E-9)(128,6.112928972911753E-7)+-(0,1.1782836729090728E-9)
                (256,5.914993622366006E-7)+-(0,1.9823982385759384E-9)(512,5.400117822896585E-7)+-(0,1.4332539966078262E-9)(1024,5.437562749750315E-7)+-(0,1.472666118740673E-9)
                (2048,5.173719938275091E-7)+-(0,1.2401981416758833E-9)(4096,5.112488987560096E-7)+-(0,4.1990737886460035E-9)(8192,5.35385331286972E-7)+-(0,3.1962640835008495E-9)
                (16384,5.100500807590444E-7)+-(0,3.474909799247706E-9)(32768,5.145859652997533E-7)+-(0,4.704702143946502E-9)(65536,5.141947682114768E-7)+-(0,5.431786011000651E-9)
                (131072,5.134673639136217E-7)+-(0,5.208857190596439E-9)(262144,5.165941866384937E-7)+-(0,5.34916671715966E-9)(524288,5.311918097224901E-7)+-(0,6.2573944961478755E-9)
                (1048576,6.030561690212624E-7)+-(0,1.1159809105498404E-8)(4194304,6.965251072020905E-7)+-(0,6.569795246647634E-9)(8388608,7.526919338009485E-7)+-(0,4.323939356735567E-9)
                (16777216,7.23890836517766E-7)+-(0,7.369014910288053E-9)
              };
    \addlegendentry{$(2+1)$ i};
    \addplot+ plot[error bars/.cd, y dir=both, y explicit]
              coordinates{
                (32,5.548583790332271E-7)+-(0,9.807415581713209E-9)(64,8.517796975824392E-7)+-(0,4.0880227247297735E-9)(128,1.2001469958170256E-6)+-(0,3.238525693537444E-9)
                (256,1.8554254051168008E-6)+-(0,1.0550931961073088E-8)(512,3.350695954723598E-6)+-(0,1.5536126371332317E-8)(1024,6.320868107854098E-6)+-(0,3.044217734857303E-8)
                (2048,1.2241815771197553E-5)+-(0,8.202891025513077E-8)(4096,2.2423325658075693E-5)+-(0,1.7006054437190767E-7)(8192,4.391009747348807E-5)+-(0,1.94822687726965E-7)
                (16384,9.466023601271866E-5)+-(0,4.152923327238261E-7)(32768,1.6573331034618307E-4)+-(0,9.460094726047847E-7)(65536,3.8080297775690145E-4)+-(0,9.729550065110115E-7)
              };
    \addlegendentry{$(10+1)$ n};
    \addplot+ plot[error bars/.cd, y dir=both, y explicit]
              coordinates{
                (32,3.1992984913943147E-6)+-(0,1.1123775007574493E-8)(64,3.0573545210517845E-6)+-(0,2.074626667748305E-8)(128,2.903931496544059E-6)+-(0,3.3910659720578954E-8)
                (256,2.712575362284579E-6)+-(0,3.694244027104919E-8)(512,2.478307737633052E-6)+-(0,4.632990922792534E-8)(1024,2.325639213345729E-6)+-(0,5.990301731714597E-8)
                (2048,2.2497261738605853E-6)+-(0,4.728706428704457E-8)(4096,2.1181110503925317E-6)+-(0,8.619151796982116E-8)(8192,2.0220032796263775E-6)+-(0,9.777544482966162E-8)
                (16384,1.978490617976351E-6)+-(0,1.2544135471235985E-7)(32768,1.8966319811350957E-6)+-(0,1.3519541231603453E-7)(65536,1.925035234927227E-6)+-(0,1.0147437475016244E-7)
                (131072,1.894476882188994E-6)+-(0,1.5418814509094415E-7)(262144,1.8511504582118127E-6)+-(0,1.7126084496846622E-7)(524288,1.8697972515272248E-6)+-(0,9.922362870076136E-8)
                (1048576,1.8795587452035955E-6)+-(0,1.7970521205133875E-7)(4194304,1.975845603927609E-6)+-(0,8.177598305417867E-8)(8388608,2.111992050182927E-6)+-(0,8.750122567283867E-8)
                (16777216,2.085938062121921E-6)+-(0,9.224937802391479E-8)
              };
    \addlegendentry{$(10+1)$ i};
\end{axis}
\end{tikzpicture}
\caption{Operation times for \textsc{OneMax}. The last letter of the algorithm's name
         means ``n'' for ``na{\"i}ve'' and ``i'' for ``incremental''.
         Means and standard deviations are shown, the latter almost always invisible.}\label{results:om}
\end{figure}

The first series of our experiments uses the \textsc{OneMax} problem with different values for the 
problem size $n$ to see how the proposed data structure works in favorable conditions. We consider
problem sizes $n=2^k$ for $k\in[5..13]$. We run each algorithm until the optimum is found, then report
the operation times as specified above. The results are presented in Fig.~\ref{results:om}.

This figure shows that the time of a single operation scales linearly with the problem size
in all algorithms using the na{\"i}ve population handling approach,
but in all algorithms using the proposed data structure it stays constant as the problem size grows.
It can be seen, however, that as the population size $\mu$ grows in the $(\mu+1)$ genetic algorithm,
the cost of an operation seems to scale linearly with the population size. This is expected, as
each operation traverses the entire population to determine the nearest neighbor. The slow increase
of the running times as the problem size reaches $10^6$ can be explained by the data structure no longer
fitting the L2 cache of the processor, which results in performance degradation by a factor
of up to three.

\subsection{Knapsack Problem, Fixed Budget}

\begin{figure}[!t]
\centering
\begin{tikzpicture}
\begin{axis}[width=0.75\columnwidth, height=0.3\textheight, legend pos=outer north east,
             xmode=log, ymode=log, cycle list name=myplotcycle, grid=both,
             xlabel={Problem size $n$}, ylabel={Average operation time, s}]
    \addplot+ plot[error bars/.cd, y dir=both, y explicit]
              coordinates{
                (32,5.934351209549526E-8)+-(0,1.0473895732853284E-10)(64,8.508587600341947E-8)+-(0,1.6192330059957932E-10)(128,1.1947011045283326E-7)+-(0,6.7685358487785415E-9)
                (256,1.882820720938967E-7)+-(0,1.4288853493925865E-10)(512,5.181025474358975E-7)+-(0,8.136826755951607E-10)(1024,8.958262599111106E-7)+-(0,1.0986530669137192E-9)
                (2048,1.893084174138528E-6)+-(0,6.909073560594772E-9)(4096,3.075468885868132E-6)+-(0,4.618479185434422E-9)(8192,1.9169225532000003E-5)+-(0,1.4233735072384763E-7)
                (16384,4.375134263200001E-5)+-(0,4.5694785109359027E-7)(32768,1.0998352652399996E-4)+-(0,3.6538789765742013E-7)(65536,3.1478902111599995E-4)+-(0,1.051365935918529E-6)
              };
    \addlegendentry{RLS n};
    \addplot+ plot[error bars/.cd, y dir=both, y explicit]
              coordinates{
                (32,1.5348578659003833E-7)+-(0,4.511817094806579E-11)(64,1.5000915140074907E-7)+-(0,6.966031375149171E-11)(128,1.5263168734438222E-7)+-(0,8.059269018838298E-11)
                (256,1.4831188777777779E-7)+-(0,6.493888549905632E-11)(512,1.5257883378469223E-7)+-(0,5.754873377731982E-11)(1024,1.5344185249042149E-7)+-(0,2.354533111805451E-11)
                (2048,1.5705758759215686E-7)+-(0,9.504858941272951E-11)(4096,1.605605950978956E-7)+-(0,1.8923561535784854E-10)(8192,1.7815095788516197E-7)+-(0,5.818274681007191E-10)
                (16384,1.9657600509803924E-7)+-(0,2.333257334829121E-10)(32768,2.429428730653381E-7)+-(0,8.261399886272838E-10)(65536,3.270937955990937E-7)+-(0,7.784581775137537E-10)
                (131072,5.065848589420615E-7)+-(0,4.191102182223145E-9)(262144,1.0392099278000002E-5)+-(0,4.441389472920399E-8)(524288,2.1659839022E-5)+-(0,2.502118193935285E-7)
                (1048576,4.2990637244E-5)+-(0,1.3539875269729608E-6)(4194304,1.9738930604000005E-4)+-(0,2.335864161854572E-6)(8388608,4.2015902554800016E-4)+-(0,1.2238638073103137E-5)
                (16777216,9.021744331039995E-4)+-(0,6.6364229422908196E-6)
              };
    \addlegendentry{RLS i};
    \addplot+ plot[error bars/.cd, y dir=both, y explicit]
              coordinates{
                (32,2.296973576813334E-7)+-(0,1.014465663184586E-9)(64,3.4619209965517236E-7)+-(0,1.0402944009745973E-9)(128,5.901861675294116E-7)+-(0,1.772904464410719E-9)
                (256,1.1576466208E-6)+-(0,2.5665021420078637E-9)(512,2.1488398098947365E-6)+-(0,6.765901769540441E-9)(1024,3.8023091760000007E-6)+-(0,1.4119774360640888E-8)
                (2048,7.4117325773333325E-6)+-(0,2.1433024431213742E-8)(4096,1.9652311953333337E-5)+-(0,6.174921405202674E-8)(8192,4.0962589424000004E-5)+-(0,3.4614514892190943E-7)
                (16384,1.0834428555199996E-4)+-(0,5.391193947497737E-7)(32768,2.2623152796800002E-4)+-(0,7.943277275269959E-7)(65536,4.964751633720001E-4)+-(0,2.1069903118257606E-6)
              };
    \addlegendentry{$(1+1)$ n};
    \addplot+ plot[error bars/.cd, y dir=both, y explicit]
              coordinates{
                (32,2.2716307117899333E-7)+-(0,1.1184763267867836E-10)(64,2.2476356011298727E-7)+-(0,1.4925166051817985E-10)(128,2.2212323737275844E-7)+-(0,6.831323676242768E-10)
                (256,2.215859804640884E-7)+-(0,1.4370429680089525E-10)(512,2.2816589588636365E-7)+-(0,1.4557474812807307E-10)(1024,2.2085404577718413E-7)+-(0,3.0788055306694034E-10)
                (2048,2.2494744200767862E-7)+-(0,4.621537062521883E-10)(4096,2.375182043786982E-7)+-(0,6.912827281671843E-11)(8192,2.650739937933915E-7)+-(0,7.671661767351158E-10)
                (16384,3.05258888618552E-7)+-(0,3.1580831451294914E-10)(32768,4.025438338210103E-7)+-(0,2.0175106377694748E-9)(65536,5.899783268397272E-7)+-(0,2.731592405040973E-9)
                (131072,9.849513921811848E-7)+-(0,8.395239570402591E-9)(262144,7.061163326666666E-6)+-(0,3.3469710394823426E-8)(524288,1.4127905181333331E-5)+-(0,1.5971325094764908E-7)
                (1048576,2.8390514066000008E-5)+-(0,5.611255008932176E-7)(4194304,1.2904448367599997E-4)+-(0,9.904780263217603E-7)(8388608,2.727446422840001E-4)+-(0,6.615005355673103E-6)
                (16777216,5.979506493439999E-4)+-(0,5.54270611483434E-6)
              };
    \addlegendentry{$(1+1)$ i};
    \addplot+ plot[error bars/.cd, y dir=both, y explicit]
              coordinates{
                (32,4.129644506391755E-7)+-(0,1.1887127775357412E-9)(64,5.541482918919767E-7)+-(0,6.838325808403163E-9)(128,8.629787854468087E-7)+-(0,2.6462840827683114E-9)
                (256,1.5811937204615382E-6)+-(0,4.1942089947101624E-9)(512,2.9071810037142856E-6)+-(0,2.203789967874603E-8)(1024,5.2578646645E-6)+-(0,2.2738095681306963E-8)
                (2048,1.2318014542000004E-5)+-(0,2.4126951964493186E-8)(4096,2.422222374400001E-5)+-(0,1.0489187457607202E-7)(8192,5.425013163200001E-5)+-(0,3.35735989856016E-7)
                (16384,1.3642686978800002E-4)+-(0,7.116181990775454E-7)(32768,3.246053178159999E-4)+-(0,1.1910087176709878E-6)(65536,5.904676561800002E-4)+-(0,1.2043786723309656E-6)
              };
    \addlegendentry{$(2+1)$ n};
    \addplot+ plot[error bars/.cd, y dir=both, y explicit]
              coordinates{
                (32,3.3142253814876035E-7)+-(0,7.233089941453369E-10)(64,3.2614711847154474E-7)+-(0,1.4946804223289738E-10)(128,3.3330639100716267E-7)+-(0,8.749465685501598E-10)
                (256,3.3689099284609035E-7)+-(0,1.2347496973567586E-9)(512,3.524869692982455E-7)+-(0,6.734931676336665E-10)(1024,3.7132568933333336E-7)+-(0,4.479885885694495E-10)
                (2048,3.898236853980584E-7)+-(0,9.49373634818621E-10)(4096,4.2490527585800655E-7)+-(0,4.812026678601304E-10)(8192,4.7420277134117644E-7)+-(0,8.797920752405586E-10)
                (16384,5.602710203771519E-7)+-(0,3.058247603147377E-9)(32768,7.002334980859044E-7)+-(0,6.351920464232056E-9)(65536,9.817382383537748E-7)+-(0,1.5407634321863127E-8)
                (131072,1.6561212669961675E-6)+-(0,6.847313923027206E-8)(262144,8.355105278399998E-6)+-(0,2.0275814480193992E-7)(524288,1.6276739898666666E-5)+-(0,1.191311906582533E-6)
                (1048576,3.2519335717999995E-5)+-(0,1.2654006590373617E-6)(4194304,1.6067422778000002E-4)+-(0,1.4911420093080468E-5)(8388608,3.2201634001600006E-4)+-(0,3.35632672831113E-5)
                (16777216,7.538344875759999E-4)+-(0,6.333358526458994E-5)
              };
    \addlegendentry{$(2+1)$ i};
    \addplot+ plot[error bars/.cd, y dir=both, y explicit]
              coordinates{
                (32,3.9651652614855063E-7)+-(0,2.137525617384806E-9)(64,6.062326620199004E-7)+-(0,1.954642901244716E-9)(128,9.631125562857145E-7)+-(0,4.038113393793142E-9)
                (256,1.7624085429565218E-6)+-(0,6.911275396372964E-9)(512,3.007466644285715E-6)+-(0,1.298179600324407E-8)(1024,6.639818990952385E-6)+-(0,5.618465428763776E-8)
                (2048,1.3133818880000004E-5)+-(0,4.688788639150789E-8)(4096,2.5377765436000012E-5)+-(0,6.757525755353148E-8)(8192,5.952742749599998E-5)+-(0,2.6953479260005754E-7)
                (16384,1.45682674144E-4)+-(0,5.007458933824654E-7)(32768,3.415081434120001E-4)+-(0,5.297572366589625E-7)(65536,6.832306142760001E-4)+-(0,1.5780184628881628E-6)
              };
    \addlegendentry{$(10+1)$ n};
    \addplot+ plot[error bars/.cd, y dir=both, y explicit]
              coordinates{
                (32,4.380195643841633E-7)+-(0,2.3665610336136762E-9)(64,4.972926840430322E-7)+-(0,5.312676455520323E-9)(128,6.204297502283216E-7)+-(0,5.325587861494516E-9)
                (256,8.955055783014496E-7)+-(0,7.121146031621103E-9)(512,1.3721556747724135E-6)+-(0,1.661223841467872E-8)(1024,2.0548949198E-6)+-(0,2.2283529590108563E-8)
                (2048,2.9488662440000006E-6)+-(0,4.1130615298920117E-8)(4096,3.699558035242424E-6)+-(0,6.198120495708589E-8)(8192,4.810129216666667E-6)+-(0,2.0117664837312666E-7)
                (16384,6.6267785515238065E-6)+-(0,2.6202560092061226E-7)(32768,9.623626033E-6)+-(0,7.020349746701503E-7)(65536,1.7176145794E-5)+-(0,3.378838922864176E-6)
                (131072,3.7181388782E-5)+-(0,7.0438525692502176E-6)(262144,6.371692081999999E-5)+-(0,1.1258239546468253E-5)(524288,1.1764041991999997E-4)+-(0,1.8912384016026353E-5)
                (1048576,2.61688898736E-4)+-(0,4.1252237657999804E-5)(4194304,0.0017699686026880009)+-(0,3.214845843207154E-4)(8388608,0.003958749689159999)+-(0,5.561251298343379E-4)
                (16777216,0.008288398597264001)+-(0,0.0012709099407040103)
              };
    \addlegendentry{$(10+1)$ i};
\end{axis}
\end{tikzpicture}
\caption{Operation times for the knapsack problem with budget of 25000 fitness evaluations.
         The last letter of the algorithm's name
         means ``n'' for ``na{\"i}ve'' and ``i'' for ``incremental''.
         Means and standard deviations are shown, the latter almost always invisible.}\label{results:knap}
\end{figure}
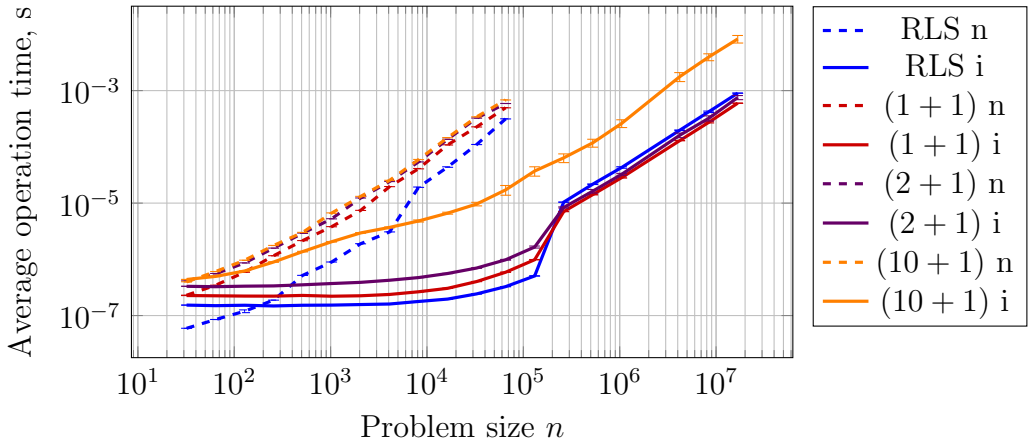

A similar experiment was conducted for the knapsack problem, where for each problem size
$n = 2^k$ for $k \in [5..13]$ a random uncorrelated instance was generated with weights and values
in the range $[10000..20000]$ and the capacity equal to half the sum of weights.
It is known that such instances are generally easy~\cite{pisinger}.
However, we did not aim at obtaining the optimum, instead, we set a computational budget of 25000 fitness
evaluations. Fig.~\ref{results:knap} shows the plots of the operation times for the considered algorithms.

In this experiment, the operation times of all the algorithms
using the na{\"i}ve population handling approach still scale linearly with the problem size,
which is not surprising. With the proposed data structure, the runtimes seem to be constant
for the algorithms with small population sizes, but only until the problem size of few thousands.
After that size, the runtimes seem to get worse until the transition around the size of $10^5$,
after which they also become linear. This, however, is easily explained by the domination of the initial
fitness evaluation: as this experiment operates with a fixed computational budget for evaluations,
for very large problem sizes computation of the first few fitness values takes more time than the
rest of the optimization.

Even despite the deficiency of the experimental setup, we can see that the $(10+1)$~GA shows
an increasing trend even for the small problem size, which, however, is significantly slower
than the one of the na{\"i}ve approach.
This can be an indication that the $(10+1)$~GA manages to maintain some diversity in the population,
which is immediately reflected in the operation times, however, the proposed data structure still
is quite efficient even in these conditions.

\subsection{Knapsack Problem, Varying Budget}

One of the possible measures of diversity in the population, which is available almost for free
in the proposed data structure, is the sum of sizes of all the patches: the larger the sum,
the bigger the diversity. Additionally, this sum may serve as an approximation of the work
each variation operator has to do, but it is unclear how good this approximation is.

We performed an additional experiment with the knapsack problem.
In this experiment, we use the $(10+1)$~GA exclusively. We use one particular knapsack problem instance
with $n=10^4$, which was randomly generated using the procedure from the previous experiment.
However, this time we record the average operation time, as well as the average patch size, every 10
evaluations until the computational budget of $10^5$ is reached.

\begin{figure}[!t]
\centering
\begin{tikzpicture}
\begin{axis}[width=0.8\columnwidth, height=0.3\textheight,
             xmode=log, ymode=log, cycle list name=myplotcycleA, grid=both,
             xlabel={Evaluations}, ylabel={Average total patch size}]
    \addplot+ table[x=evals, y=size-avg] \sizebyevalstable;
    \addplot+ [name path=maxP] table[x=evals, y=size-max] \sizebyevalstable;
    \addplot+ [name path=minP] table[x=evals, y=size-min] \sizebyevalstable;
    \addplot [blue!20!white] fill between [of=minP and maxP, on layer=axis background];
\end{axis}
\end{tikzpicture}
\caption{Average patch sizes for the $(10+1)$~GA on the knapsack problem with $n=10^4$
         as a function of the number of fitness evaluations. Means, minima and maxima are shown.}
         \label{results:knap:div:size}
\vspace{3ex}
\begin{tikzpicture}
\begin{axis}[width=0.8\columnwidth, height=0.3\textheight,
             xmode=log, ymode=log, cycle list name=myplotcycleB, grid=both,
             xlabel={Evaluations}, ylabel={Average operation time, s}]
    \addplot+ table[x=evals, y=time-avg] \timebyevalstable;
    \addplot+ [name path=maxP] table[x=evals, y=time-max] \timebyevalstable;
    \addplot+ [name path=minP] table[x=evals, y=time-min] \timebyevalstable;
    \addplot [blue!20!white] fill between [of=minP and maxP, on layer=axis background];
\end{axis}
\end{tikzpicture}
\caption{Average operation times for the $(10+1)$~GA on the knapsack problem with $n=10^4$
         as a function of the number of fitness evaluations. Means, minima and maxima are shown.}
         \label{results:knap:div:time}
\end{figure}

The plot of average patch sizes against the number of evaluations is presented
in Fig.~\ref{results:knap:div:size}, and the plot of average operation times
is presented in Fig.~\ref{results:knap:div:time}. Both plots are somewhat similar:
in the beginning, both quantities quickly reach the maximum value, then they drop to the small
values, and starting from roughly 600 fitness evaluations they decrease very slowly. The initial growth
phase seems somewhat unexpected, but it can be explained by a number of individuals which did not
survive, but have not yet been deleted from the tree. The quick decrease corresponds to the gradual
loss of diversity, which is not unexpected.

What is interesting, however, is that the patch size does not drop to values of 0 or 1 for a long time,
which indicates that a noticeable amount of diversity is still present in the population.
For up to $10^4$ evaluations, the average total patch size is greater than 30 for a population of 10,
which cannot happen if the individuals are set aside by single bit flips. It can happen though that
there are just two different individuals located at a large distance, with the rest of the population
being their copies. However, this can still result in different population dynamics compared to the
population-less algorithms.

\begin{figure}[!t]
\centering
\begin{tikzpicture}
\begin{axis}[width=0.8\columnwidth, height=0.3\textheight,
             xmode=log, ymode=log, cycle list name=myplotcycleA, grid=both,
             xlabel={Average total patch size}, ylabel={Average operation time, s}]
    \addplot+ table[x=size, y=time-avg] \timebysizetable;
    \addplot+ [name path=maxP] table[x=size, y=time-max] \timebysizetable;
    \addplot+ [name path=minP] table[x=size, y=time-min] \timebysizetable;
    \addplot [blue!20!white] fill between [of=minP and maxP, on layer=axis background];
\end{axis}
\end{tikzpicture}
\caption{Average operation times for the $(10+1)$~GA on the knapsack problem with $n=10^4$
         as a function of the average patch size. Means, minima and maxima are shown.}
         \label{results:knap:div:corr}
\end{figure}
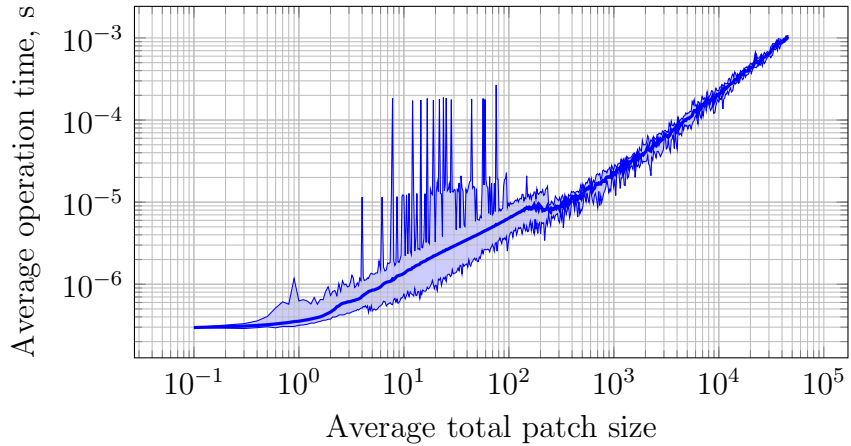

Since Fig.~\ref{results:knap:div:size} and~\ref{results:knap:div:time} look very similar, one may
expect that the total patch size and the average operation time are strongly correlated, which
also matches the employed algorithms well. Fig.~\ref{results:knap:div:corr} plots the average operation time
as a function of the average total patch size. While the correlation is indeed high (the Pearson
correlation coefficient was 0.9327), Fig.~\ref{results:knap:div:corr} suggests that the dependency
might be more complicated. Though the sublinear parts of the employed algorithms
can affect the linearity for very small patch sizes, the explanation of the abrupt change of the trends
around the patch size of 200 may require additional investigations.


\section{Conclusion and Future Work}
\label{sec:conclusion}

We proposed a data structure for storing populations, the minimum spanning tree of patches,
that has a good potential of improving time and memory complexity of population-based genetic
algorithms in a similar way that is already possible for mutation-only algorithms
and gray-box optimizers.

\textbf{Different search spaces}. Most if not all of the proposed ideas can be implemented
in discrete search spaces other than bit strings. Care should be taken, however, of how
exactly the patches are combined, and especially reversed. Permutations already present a challenge
there, as a related paper shows~\cite{bassinB-gecco20-oll-perm}.

\textbf{Hash functions for non-revisiting algorithms}. Polynomial hash functions can be incrementally
computed in a way similar to fitness functions. This can be used to maintain a hash table
of all individuals in the population and prevent sampling the same individual more than once. Note that
a patch of size zero will otherwise be created for such a duplicate, however, filtering these out based
on hash tables is more efficient.

\textbf{Not only trees to make paths shorter}. The minimality of the minimum spanning tree can be,
to some extent, sacrificed for making the average paths between individuals shorter. It is not clear
what is the most efficient design of such a graph, but most likely the number of edges should still be
linear in the number of vertices.

\textbf{More than one complete individual}. If the population size is large enough, or if there is
significant diversity in the population, the total size of all the patches may be of the same order,
or more, as the size of an individual. The data structure may then benefit from using more than one
complete individual: a request to perform an operator on an individual may be fulfilled by, for instance,
the closest complete individual. The proper management of the positioning of the complete individuals
may be tricky and should be a subject of further investigation.

\textbf{Large populations}. It is undesirable to traverse the entire population in the case it
becomes large. Advanced balanced trees and path handling data structures, such as
splay trees~\cite{splay-tree} and link-cut trees~\cite{link-cut-tree}, have a potential to
improve the performance further. They may also help in a better, non-lazy, handling of discarding
individuals by treating the minimum spanning tree as a dynamic structure. With more effort,
it is probably possible to implement specialized queries on large populations,
such as searching for individuals at a certain distance faster than by complete enumeration.
With an efficient implementation, this may go as far as remembering all the individuals sampled througout
the search process.

\textbf{Higher arities}. Similarly to supporting crossovers, it is possible to support higher-arity
operators, for instance ternary operators common to differential
evolution~\cite{doerrZ-binary-differential-evolution,differential-evolution}.
This requires only a moderate modification of the data structure behind the mutable patch,
which would still occupy the linear amount of memory.

\textbf{Na{\"i}ve bootstrapping}. Finally, the proposed data structure can be used not from the very
start of the algorithm, but once the diversity decreases somewhat, which would alleviate the problem 
of large overheads during the first iterations of the algorithms found in the experiments.

\bibliographystyle{abbrv}
\bibliography{../../../../bibliography}

\end{document}